\documentclass{article}

\usepackage{arxiv}

\usepackage[utf8]{inputenc} 
\usepackage[T1]{fontenc}    
\usepackage{hyperref}       
\usepackage{url}            
\usepackage{booktabs}       
\usepackage{amsfonts}       
\usepackage{nicefrac}       
\usepackage{microtype}      
\usepackage{amsmath}
\usepackage{amssymb}
\usepackage{cleveref}       
\usepackage{lipsum}         
\usepackage{graphicx}
\usepackage{natbib}
\usepackage{doi}
\usepackage{multirow}
\usepackage{lineno}

\title{Structure-Guided Diffusion Model for EEG-Based Visual Cognition Reconstruction}

\author{ {\hspace{1mm}Yongxiang Lian}\thanks{These authors contributed equally to this work and are co-first authors.} \\
    Department of Automation\\
    Tsinghua University\\ 
    Haidian District, 100084, Beijing, China\\
	\texttt{lianyx20@mails.tsinghua.edu.cn} \\
    \And
	{\hspace{1mm}Yueyang Cang$^*$} \\
	Department of Automation\\
    Tsinghua University\\ 
    Haidian District, 100084, Beijing, China\\
	\texttt{cangyy23@mails.tsinghua.edu.cn} \\
    \And
	{\hspace{1mm}Pingge Hu} \\
	China Academy of Information and\\ Communications Technology\\ 
    Haidian District, 100084, Beijing, China\\
	\texttt{hupingge@caict.ac.cn} \\
    \And
	{\hspace{1mm}Yuchen He} \\
	Department of Automation\\
    Tsinghua University\\ 
    Haidian District, 100084, Beijing, China\\
	\texttt{heyuchen25@mails.tsinghua.edu.cn} \\
	\And
	\href{https://orcid.org/0000-0001-5204-9461}{\includegraphics[scale=0.06]{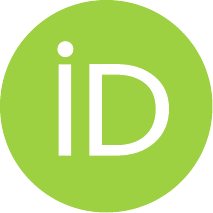}\hspace{1mm}Li Shi} \\
	Department of Automation\\
    Tsinghua University\\ 
    Haidian District, 100084, Beijing, China\\
	\texttt{shilits@mail.tsinghua.edu.cn} \\
}

\renewcommand{\shorttitle}{\textit{arXiv} Template}

\hypersetup{
pdftitle={Structure-Guided Diffusion Model for EEG-Based Visual Cognition Reconstruction},
pdfsubject={q-bio.NC, cs.LG},
pdfauthor={Yongxiang Lian, Yueyang Cang, Pingge Hu, Yuchen He, Li Shi},
pdfkeywords={Electroencephalography, Diffusion model, Structure guided, Cognition reconstruction},
}

\begin{document}
\maketitle

\begin{abstract}
Objective:
Decoding visual information from electroencephalography (EEG) is an important problem in neuroscience and brain–computer interface (BCI) research. Existing methods are largely restricted to natural images and categorical representations, with limited capacity to capture structural features and to differentiate objective perception from subjective cognition. We propose a Structure-Guided Diffusion Model (SGDM) that incorporates explicit structural information for EEG-based visual reconstruction.
Approach:
SGDM is evaluated on the Kilogram abstract visual object dataset and the THINGS natural image dataset using a two-stage generative mechanism. The framework combines a structurally supervised variational autoencoder with a spatiotemporal EEG encoder aligned to a visual embedding space via contrastive learning. Structural information is integrated into a diffusion model through ControlNet to guide image generation from EEG features.
Results:
SGDM outperforms existing methods on both abstract and natural image datasets. Reconstructed images achieve higher fidelity in low-level visual features and semantic representations, indicating improved decoding accuracy and strong generalization across diverse visual domains. Spatiotemporal analysis of EEG signals further reveals hierarchical structural encoding patterns, consistent with the neural dynamics of visual cognition.
Significance:
These findings validate the effectiveness of SGDM in capturing explicit structural geometry and generating images with high fidelity to individual cognitive representations. By enabling decoding of complex visual content from EEG signals, the framework extends neural decoding beyond low-dimensional or categorical outputs. This supports BCIs with increased degrees of freedom for intention decoding and more flexible brain-to-machine communication.
\end{abstract}


\section{Introduction}
Decoding human thoughts from brain activity has long been regarded as one of the central goals of neuroscience~\cite{he2018neural,murad2024unveiling,tong2012decoding}. In recent years, the rapid development of artificial intelligence has made it increasingly feasible to decode neural representations of visual perception and cognition~\cite{chen2024toward,scotti2023reconstructing,wang2025progress}. Among various neuroimaging techniques, electroencephalography (EEG) offers millisecond-level temporal resolution that captures subtle dynamics of brain activity~\cite{luck2014introduction,wang2025progress}. Due to its non-invasiveness and accessibility, EEG has become a crucial tool for investigating visual perception and cognitive processes~\cite{cheng2025fine,zeng2023dm}. However, the low signal-to-noise ratio, significant inter-individual variability, and complex associations among hierarchical visual semantics make it challenging to reconstruct images that humans “see” from EEG signals~\cite{zeng2023dm,singh2024learning}. Furthermore, generating images that reflect subjective cognition introduces additional challenges related to experimental paradigms and generative modeling mechanisms~\cite{palazzo2020decoding,li2024visual}.

In recent years, researchers have attempted to extract visually relevant representations from EEG signals using deep learning models~\cite{spampinato2017deep,lawhern2018eegnet}, aiming to decode perceptual images evoked by visual stimuli. Early studies employed convolutional neural networks (CNNs) and recurrent neural networks (RNNs) to achieve relatively accurate object category recognition~\cite{guo2024investigating,bashivan2016learning,roy2019deep}. Subsequently, the integration of contrastive learning and cross-modal embedding techniques enabled the alignment of EEG signals with visual semantic spaces~\cite{radford2021clip,jia2021scaling}. Combined with the progress in diffusion-based generative models~\cite{ho2020ddpm,rombach2022ldm,saharia2022imagen}, these advances have significantly propelled image retrieval and reconstruction from neural signals. In particular, the remarkable reconstruction results achieved using fMRI have provided substantial impetus to the development of this field~\cite{takagi2023high}.

Inspired by the hierarchical processing mechanism of the human visual system~\cite{felleman1991distributed,kriegeskorte2008representational,yamins2016using}, recent studies have begun to emphasize hierarchical visual representations that go beyond categorical semantics~\cite{beliy2019voxel,kar2019evidence}. By incorporating multi-level features into the latent space of diffusion models, these approaches aim to enhance reconstruction fidelity~\cite{rombach2022ldm,takagi2023high}. However, decoding low-level visual features from EEG remains particularly challenging~\cite{roy2019deep,palazzo2020decoding}. Existing approaches often align EEG features with high-dimensional CLIP embeddings~\cite{radford2021clip,jia2021scaling} or apply joint training strategies to achieve cross-modal mapping~\cite{yang2022crossmodal,li2024visual}. Yet, due to the substantial distributional discrepancy between EEG and visual semantic spaces, these methods typically ensure similarity only at a numerical level, leading to unstable training, overfitting, and poor generalization across subjects~\cite{palazzo2020decoding,spampinato2017deep}. Consequently, both structural accuracy and semantic consistency remain limited~\cite{shen2019endtoend,zeng2023dm}.

In summary, current EEG-based visual decoding approaches suffer from three major limitations~\cite{roy2019deep,cheng2025fine}:
(1) Over-reliance on categorical representations. Existing models are predominantly based on natural image stimuli and heavily depend on categorical discrimination, while overlooking structured representations of visual objects at the cognitive level, resulting in poor generalization under complex cognitive scenarios~\cite{singh2024learning};
(2) Lack of explicit hierarchical structural information. Most studies focus on low-level visual features such as category, color, and contour, without datasets or modeling frameworks capable of explicitly describing local-to-global structural relationships~\cite{zeng2023dm};
(3) Difficulty in aligning individual cognitive interpretations. Current experimental paradigms primarily focus on objective "seen" stimuli and often overlook the subjective cognitive processes involved in how individuals decompose and interpret visual structures. This neglect leads to a "semantic gap" between generic neural signals and the fine-grained cognitive representations of specific objects, hindering a deeper understanding of how the brain encodes complex visual hierarchies~\cite{li2024visual}.

To address these challenges, this study proposes a Structure-Guided Diffusion Model (SGDM) for EEG-based visual cognition reconstruction~\cite{li2024visual,rombach2022ldm}. Instead of solely relying on categorical labels, the model leverages detailed human-annotated structural priors to bridge the gap between low-level visual perception and high-level cognitive interpretation~\cite{zhang2023controlnet,ye2023ipadapter}. We utilize the Kilogram dataset, which comprises abstract visual objects with subjectively defined local-to-global structures. This allows us to model how the brain represents organized visual entities through the lens of individual cognitive consensus, rather than simple stimulus-response mapping.

Based on this foundation, the SGDM framework incorporates three key mechanisms:

1. Structured priors and explicit prediction: A variational autoencoder (VAE) is employed to learn structural encodings of visual objects, enabling generative prediction from EEG latent representations to structural embeddings~\cite{rombach2022ldm,kar2019evidence};

2. Cross-modal semantic alignment via multi-semantic annotations: To capture the cognitive diversity of visual objects, we fine-tune a CLIP model using abstract images paired with detailed, human-labeled structural descriptions. Contrastive learning is then applied to align EEG features with these cognitively enriched visual embeddings~\cite{radford2021clip,jia2021scaling,frome2013devise}, ensuring the model decodes not just a silhouette, but the intended structural meaning;

3. Controllable diffusion generation: Structural predictions are embedded into the ControlNet diffusion process, allowing cognitively generated images to achieve structural controllability while maintaining semantic individuality~\cite{zhang2023controlnet,ye2023ipadapter}.

Experimental results on the Kilogram and THINGS datasets demonstrate that SGDM significantly outperforms existing methods in structural prediction accuracy, image generation quality, and semantic consistency~\cite{li2024visual,takagi2023high}. Moreover, spatiotemporal feature analysis reveals that the structural encoding patterns learned by the model exhibit strong correspondence with neural mechanisms underlying visual cognition~\cite{felleman1991distributed,kar2019evidence}. Overall, this work introduces a new structured decoding paradigm for EEG-based visual cognition reconstruction and provides fresh insights into the cognitive guidance mechanisms of multimodal generative models.

\section{Methods}
\begin{figure*}[htpb]
    \centering
    \includegraphics[width=0.80\linewidth]{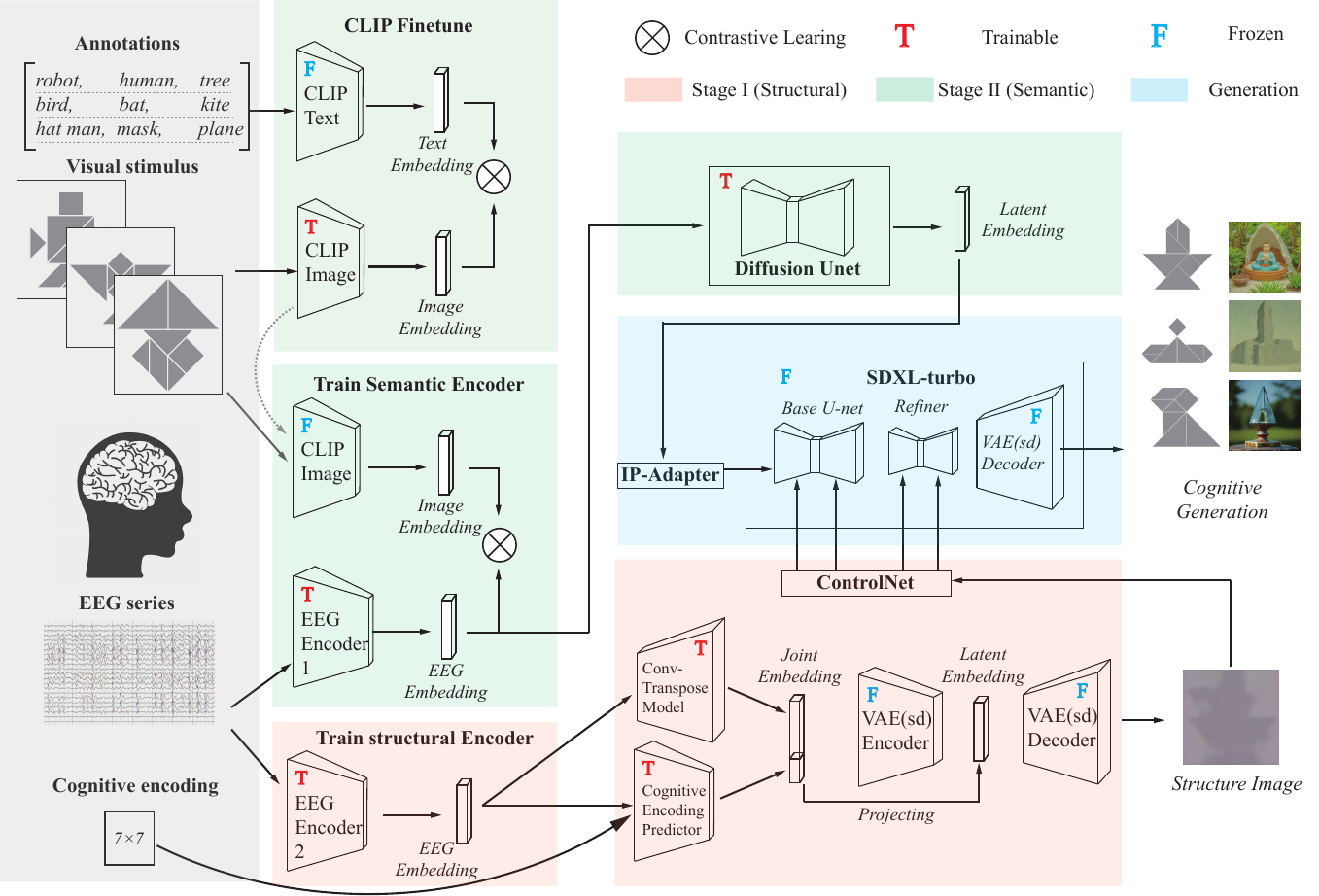}
    \caption{
        Overview of the proposed \textbf{Structure-Guided Diffusion Model (SGDM)} framework. 
        The system consists of four main modules: 
        (1) a CLIP text–image dual encoder~\cite{radford2021clip} for unified semantic representation learning, 
        (2) an EEG semantic encoder trained to align with the CLIP image space~\cite{li2024visual}, 
        (3) a structure prediction module based on the SDXL-turbo VAE~\cite{rombach2022ldm} for generating spatial priors, 
        and (4) a dual-constrained diffusion generator that integrates semantic and structural conditions 
        via IP-Adapter~\cite{ye2023ipadapter} and ControlNet~\cite{zhang2023controlnet} 
        to synthesize cognitively consistent images.
    }
    \label{fig:sgdm_pipeline}
\end{figure*}

In this section, we present the proposed \textbf{Structure-Guided Diffusion Model (SGDM)}(Fig.~\ref{fig:sgdm_pipeline}), 
a unified framework for EEG-based visual cognition reconstruction. 
Given raw EEG signals as input, SGDM aims to explicitly disentangle structural information 
and establish cross-modal correlations between \textit{EEG–Vision–Text}, 
ultimately generating visual images consistent with individual cognitive representations.

SGDM consists of four collaborative modules: 
(1) a CLIP-based text–image dual encoder to construct a unified semantic space, 
(2) an EEG semantic encoder to extract and align high-level EEG features, 
(3) a structure prediction module based on SDXL-turbo VAE for generating visual structure maps, 
and (4) a dual-constrained diffusion generator for semantic–structural guided image synthesis.

\subsection{Dataset and Preprocessing}
\subsubsection{Dataset}

The experiments were conducted on two complementary datasets: the Kilogram Abstract Visual Object Dataset and the THINGS Natural Image Dataset.

Kilogram~\cite{ji2022abstract} contains over 900 tangrams, each of which was annotated with at least 10 segmentation annotations (See Supplementary data). On the basis of this dataset, we constructed the Cognitive Encoding to establish a mathematical foundation for quantification and to provide supervisory information. We further established quantitative indices of tangram stimuli, including abstraction level and local feature density. Based on these measures, 75 representative tangram figures were selected, and EEG experiments\cite{lian2025multidimensional} were conducted with 24 participants with Psychopy toolbox~\cite{peirce2019psychopy2}.
The THINGS dataset~\cite{gifford2022large} contains 1,854 natural object categories with 26,000 high-resolution images, serving as a perceptual baseline for generalization and cross-domain evaluation.

\subsubsection{EEG Acquisition and Preprocessing}

The EEG experimental data were obtained from our previous work \cite{lian2025multidimensional}. Detailed information regarding \textbf{visual stimuli, hardware specifications, experimental parameters, preprocessing pipelines, electrode configurations, and ocular artifact removal} is provided in the \textbf{Supplementary Data}.

\begin{figure}[htpb]
    \centering
    \includegraphics[width=0.5\linewidth]{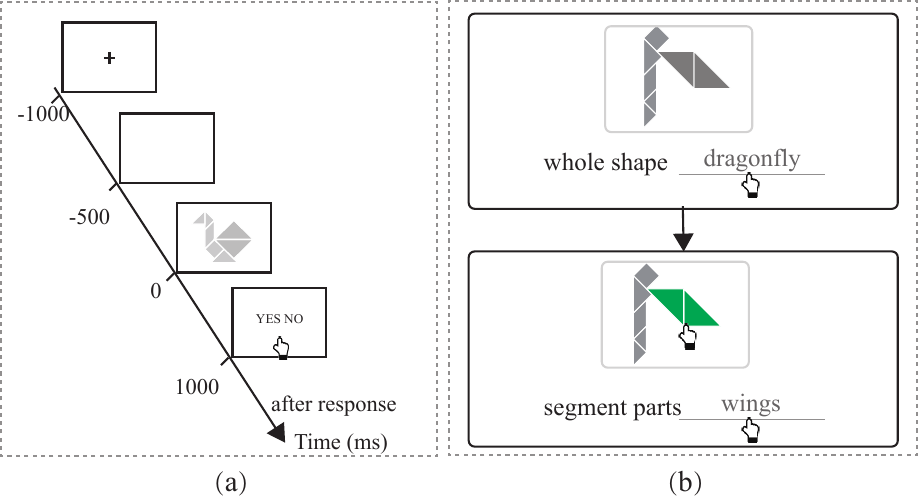}
    \caption{
        EEG data acquisition and labeling tasks on the Tangram dataset.
        (a)Experimental paradigm for Tangram EEG data acquisition.
        (b)Global annotation and semantic segmentation tasks for Tangram images.
    }
    \label{fig:eegexp_pipeline}
\end{figure}

The experimental paradigm is illustrated in Fig.~\ref{fig:eegexp_pipeline}. Participants sequentially completed a visual cognition EEG acquisition task (Fig.~\ref{fig:eegexp_pipeline}(a)) and a Tangram labeling task (Fig.~\ref{fig:eegexp_pipeline}(b)). In the EEG acquisition task, each trial began with a fixation cross presented for 500 ms, followed by a 500 ms blank screen. At 0 ms, a visual stimulus was randomly presented for a duration of 1000 ms. Upon stimulus offset, a response prompt appeared, requiring participants to perform a categorical judgment (i.e., whether the object was an animal).
To eliminate motor effector bias from the left and right hands, the mapping between the response keys and the "Yes/No" options was randomized for each trial. The trial was terminated upon the participant's key press. An inter-trial interval (ITI) of 1.5--2.5 s was applied to mitigate expectancy effects.

Following the EEG acquisition, a behavioral labeling experiment was conducted. Participants provided detailed English annotations for the stimulus images presented during the previous session. The labeling process consisted of two distinct sub-tasks: global annotation and semantic segmentation. These annotated data represent the participants' subjective cognitive perception of the Tangram stimuli.

EEG preprocessing followed the standardized MNE \cite{gramfort2013meg} protocol described in our previous study \cite{lian2025multidimensional} (see \textbf{Supplementary Data} for details). Unless otherwise specified, all preprocessing parameters and channel configurations remained consistent with the established protocol \cite{lian2025multidimensional}, ensuring cross-subject consistency and signal reliability.

\subsubsection{Data Splitting and Leakage Avoidance}

The dataset was partitioned into training, validation, and test sets using an $8:1:1$ ratio. The partitioning was performed at the image level rather than the subject level to ensure that the visual stimuli in the test set were entirely absent from the training set. This image-wise isolation protocol was strictly followed to prevent data leakage, thereby avoiding overestimation of the model's performance.

\subsubsection{Statistical Analysis}

Group-level differences between reconstruction methods were evaluated using two-tailed paired t-tests unless otherwise specified. p-values were adjusted for multiple comparisons using the Benjamini–Hochberg false discovery rate (FDR) correction. Statistical significance was defined at $p < 0.05$. Bootstrapped confidence intervals (1,000 resamples) were computed to estimate inter-subject variability.

\subsection{CLIP Text–Image Encoder}

\begin{figure}[htpb]
    \centering
    \includegraphics[width=0.5\linewidth]{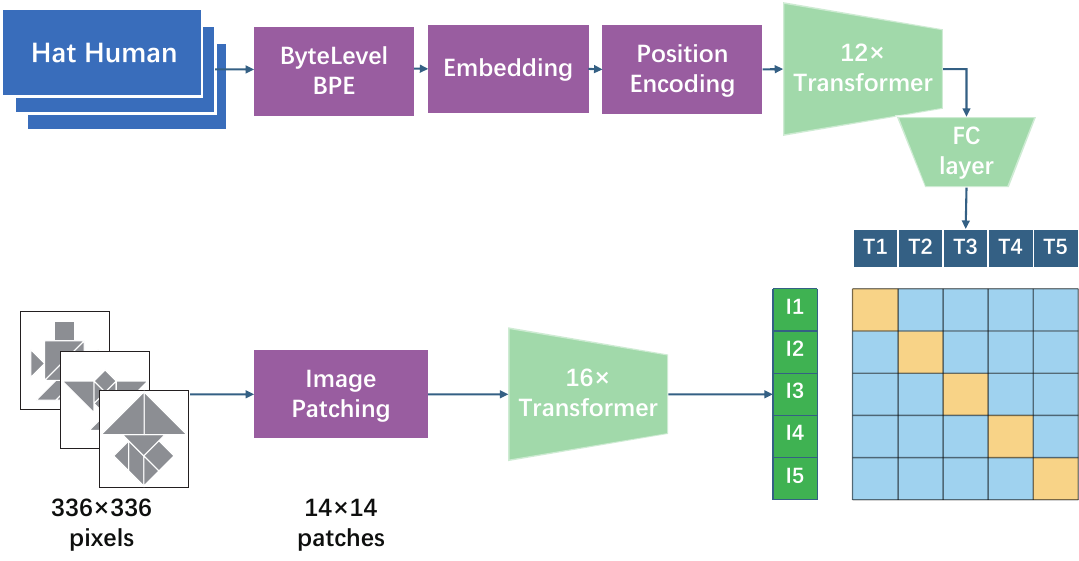}
    \caption{
         Overview of the CLIP-based text–image dual encoder~\cite{radford2021clip}. 
        The text branch encodes tokenized descriptions through Transformer layers, 
        while the image branch processes visual patches with a Vision Transformer. 
        Both outputs are projected into a shared 1024-dimensional semantic space.
    }
    \label{fig:clip}
\end{figure}

To establish a consistent cross-modal reference for EEG–vision alignment, 
we first construct a unified semantic space between natural language and visual stimuli. 
Specifically, we adopt the CLIP ViT-H/14 architecture~\cite{radford2021clip} 
as the foundation for cross-modal representation learning. 
The pretrained text–image dual encoders provide robust semantic alignment 
and serve as a semantic baseline for EEG-to-vision mapping. 
An overview of the dual-encoder architecture is illustrated in Fig.~\ref{fig:clip}.

\subsubsection{Text Encoder}

The text encoder transforms natural-language annotations into normalized semantic embeddings. 
It consists of a 12-layer Transformer with a hidden dimension of 768 and 12 attention heads. 
Each input text sequence is tokenized using CLIP’s ByteLevel BPE tokenizer, 
padded or truncated to a fixed length of 77 tokens, and enriched with \texttt{<start>} and \texttt{<end>} symbols. 
The resulting embeddings are mapped to 768 dimensions and combined with learnable positional encodings, 
forming an input tensor of shape $(B, 77, 768)$. 
After Transformer encoding, the hidden state corresponding to the \texttt{<end>} token is linearly projected into 
a 1024-dimensional semantic vector:
\[
z_T = \mathrm{Proj}(h_{\texttt{<end>}})
\]
which is directly comparable to the CLIP image encoder output. 
The 1024-dimensional embedding design ensures compatibility with the EEG latent space in later alignment modules.

\subsubsection{Image Encoder}

The image encoder follows the ViT-H/14 backbone, consisting of 32 Transformer blocks with a hidden size of 1280. 
Input images are resized to $224\times224$, divided into non-overlapping patches, linearly projected, 
and augmented with positional encodings. 
After Transformer encoding, the [CLS] token is projected to a 1024-dimensional embedding:
\[
z_I = \mathrm{Proj}(h_{\text{CLS}})
\]
Since CLIP is pretrained primarily on natural images and lacks abstraction-awareness, 
we fine-tune only the last three Transformer layers and the projection head using the InfoNCE contrastive loss:
\[
\mathcal{L}_{\mathrm{CLIP}} = -\frac{1}{B} \sum_i 
\log \frac{\exp(\mathrm{sim}(z_T^{(i)}, z_I^{(i)}) / \tau)}
{\sum_j \exp(\mathrm{sim}(z_T^{(i)}, z_I^{(j)}) / \tau)}
\]
where $\tau$ denotes the temperature parameter controlling similarity distribution sharpness. 
The resulting 1024-D embeddings define a unified semantic space that serves as the foundation 
for EEG–vision–text alignment in subsequent modules.

\subsection{EEG–Semantic Alignment with Diffusion Image Embedding}

This module aims to bridge EEG representations and visual semantics in the CLIP latent space, 
providing high-level priors for downstream image generation. 
It consists of two cooperative components: 
(1) an \textbf{EEG semantic encoder} that extracts and aligns cognitive representations from EEG signals, 
and (2) a \textbf{Prior Diffusion Model} that transforms EEG-derived embeddings into CLIP-like latent distributions, 
ensuring consistency with visual semantic priors.

\subsubsection{EEG Encoder (ATM)}

\begin{figure}[htpb]
    \centering
    \includegraphics[width=0.5\linewidth]{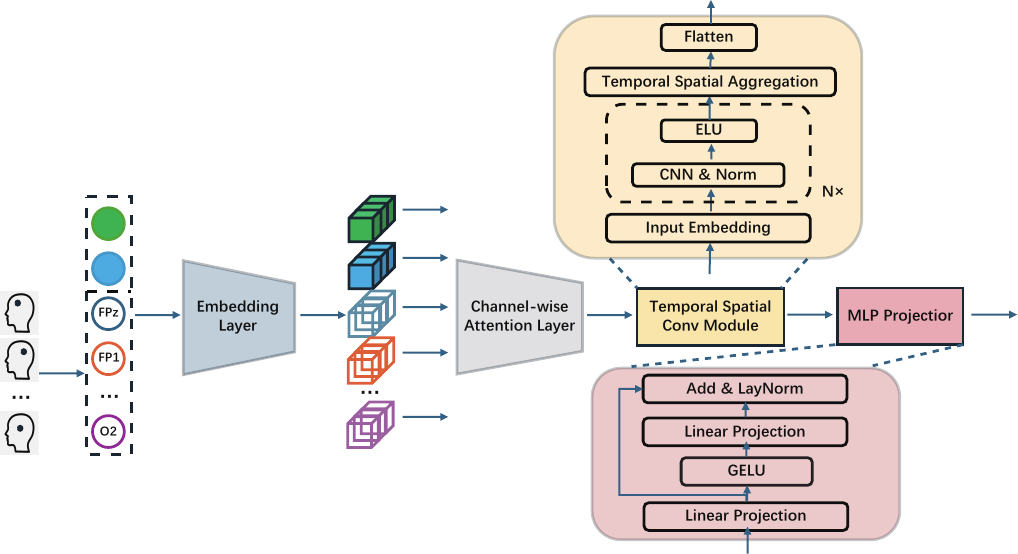}
    \caption{
        Architecture of the \textbf{Adaptive Thinking Mapper (ATM)} EEG encoder~\cite{li2024visual}. 
        Raw EEG signals from multiple channels are first embedded through an embedding layer, 
        followed by a channel-wise attention mechanism to model inter-channel dependencies. 
        The temporal–spatial convolution module aggregates local and global EEG dynamics, 
        and the MLP projector with residual and normalization layers 
        maps the extracted features into the 1024-dimensional CLIP semantic space.
    }
    \label{fig:atm}
\end{figure}

The EEG encoder follows the architecture of the \textbf{Adaptive Thinking Mapper (ATM)}~\cite{li2024visual}, 
which is designed to extract semantic representations from EEG signals 
and align them with the CLIP embedding space. 
This design is inspired by previous EEG-to-vision alignment work~\cite{spampinato2017deep}.
Given EEG inputs $\mathbf{E} = [e_1, e_2, \ldots, e_C]^\top \in \mathbb{R}^{C\times T}$, 
the encoder maps them into the CLIP semantic space:
\[
\mathbf{Z}_E = f_{\text{EEG}}(\mathbf{E}; \theta_E) \in \mathbb{R}^{B\times1024}
\]
The encoder integrates \textit{channel-wise attention}, \textit{temporal–spatial convolution}, 
and an \textit{MLP projector with residual blocks and LayerNorm} 
to capture rich temporal–spatial dependencies within EEG signals. 
The structure of the ATM encoder is illustrated in Fig.~\ref{fig:atm}.

\textit{(1) Channel-wise Attention.}
Temporal embeddings are first computed via 1D convolution and positional encoding:
\[
H_c = \mathrm{Conv1D}(e_c) + P_t
\]
Attention weights are then obtained through a self-attention operation:
\[
\widetilde{\mathbf{H}} = 
\mathrm{softmax}\!\left(
\frac{(\mathbf{H}W_Q)(\mathbf{H}W_K)^\top}{\sqrt{d_k}}
\right)(\mathbf{H}W_V),
\quad
\mathbf{H}' = \widetilde{\mathbf{H}} + \mathbf{H}
\]

\textit{(2) Temporal–Spatial Convolution.}
The model aggregates features across temporal and spatial dimensions:
\[
\mathbf{F} = \sigma(W_t * \mathbf{H}' + W_s * \mathbf{H}' + b)
\]
where $*$ denotes convolution and $\sigma(\cdot)$ is the ReLU activation function.

\textit{(3) MLP Projector.}
Flattened features are passed through a two-layer MLP with residual and normalization layers 
to produce semantic embeddings aligned with the CLIP image space:
\[
\mathbf{Z}_E = W_2 \phi(W_1 \mathrm{Flatten}(\mathbf{F}) + b_1) + b_2
\]

\textit{(4) Training Objective.}
The encoder is optimized with a joint contrastive–regression objective:
\[
\mathcal{L}_{\text{sem}} = 
\lambda_1 \mathcal{L}_{\text{CLIP}} + 
\lambda_2 \mathcal{L}_{\text{MSE}},
\quad
\mathcal{L}_{\text{MSE}} = \|\mathbf{Z}_E - z_I\|_2^2
\]
where $\mathcal{L}_{\text{CLIP}}$ is the contrastive loss defined in the previous subsection.

\subsubsection{Prior Diffusion}

To further align EEG-derived embeddings with the visual semantic manifold, 
a conditional diffusion prior~\cite{ho2020ddpm} is introduced to model the distribution $p(z_I \mid \mathbf{Z}_E)$. 
The model learns to predict Gaussian noise in the CLIP latent space:
\[
z_I^t = z_I + \sigma_t \epsilon, \quad 
\hat{\epsilon} = \epsilon_{\text{prior}}(z_I^t, t, \mathbf{Z}_E)
\]
and minimizes the denoising objective:
\[
\mathcal{L}_{\text{prior}} =
\mathbb{E}_{t, z_I, \epsilon} 
\big[\|\epsilon_{\text{prior}}(z_I^t, t, \mathbf{Z}_E) - \epsilon\|_2^2\big]
\]
This probabilistic alignment enables EEG features to be transformed into CLIP-consistent semantic latents, 
providing a robust prior for downstream visual reconstruction.

\subsection{Structure Prediction Module}

To provide spatial priors for visual reconstruction, 
we introduce an EEG-based \textbf{Structure Prediction Module} 
that estimates coarse visual structures from neural signals. 
This module takes EEG features as input and generates a low-resolution structural representation, 
which guides the subsequent diffusion-based image synthesis.

\subsubsection{Network Architecture}

\begin{figure}[htpb]
    \centering
    \includegraphics[width=0.5\linewidth]{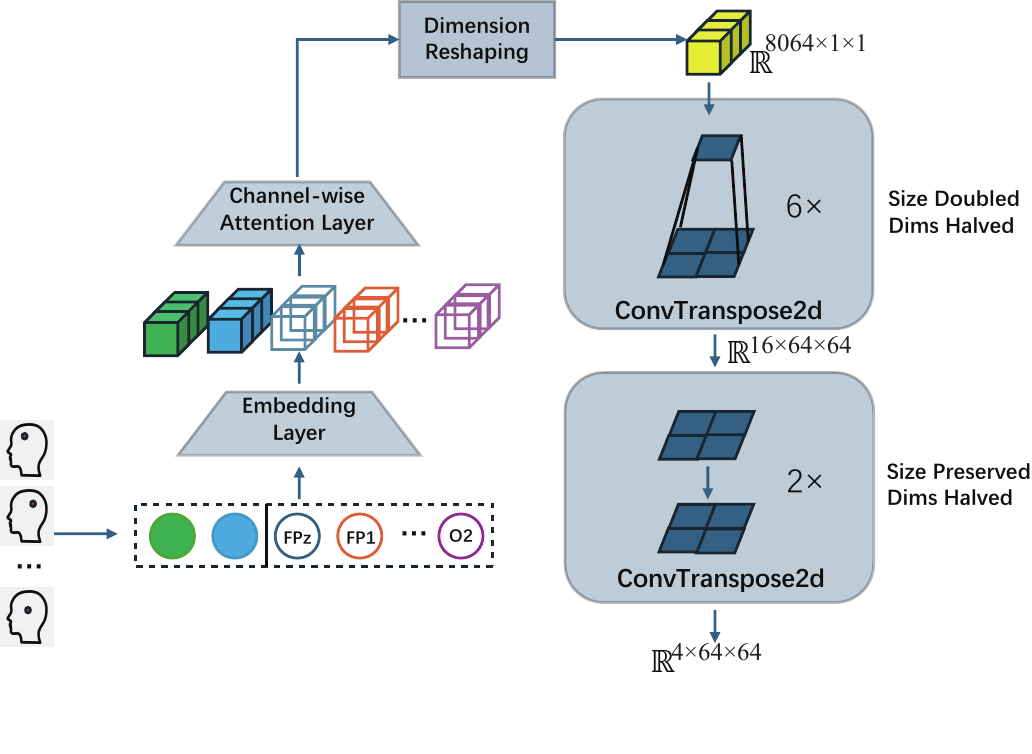}
    \caption{
        Architecture of the EEG-based \textbf{Structure Prediction Module}. 
        EEG signals are embedded and processed through a channel-wise attention layer 
        to capture spatiotemporal dependencies. 
        The resulting latent representations are reshaped and passed through stacked 
        transposed convolutional layers to produce a coarse visual structure map, 
        providing spatial priors for the subsequent diffusion generator.
    }
    \label{fig:structure_encoder}
\end{figure}

As illustrated in Fig.~\ref{fig:structure_encoder}, 
the network follows an encoder–decoder paradigm similar to ControlNet~\cite{zhang2023controlnet}. 
Raw EEG signals are first processed through an \textit{embedding layer} 
and a \textit{channel-wise attention layer} to obtain discriminative spatiotemporal representations. 
These features are then reshaped into a compact latent tensor and progressively upsampled via multiple 
\textit{ConvTranspose2d} layers to generate structural feature maps. 
The upper block contains six transposed convolutional layers that double the spatial size 
and halve the feature dimensions, 
while the lower block contains two transposed convolutional layers that preserve the resolution 
and refine the generated structure. 
The resulting feature maps $\mathbf{S} \in \mathbb{R}^{4\times64\times64}$ 
serve as the visual structural prior for diffusion generation.

\subsubsection{Training Objective}

The structure prediction network is trained to minimize the mean squared error 
between the predicted structure map $\mathbf{S}$ and the reference structural map $\mathbf{S}_{\text{gt}}$ 
extracted from the SDXL-turbo VAE encoder~\cite{rombach2022ldm}:
\[
\mathcal{L}_{\text{struct}} = \|\mathbf{S} - \mathbf{S}_{\text{gt}}\|_2^2
\]
This structural supervision ensures that the predicted maps capture 
the global spatial layout and object contours consistent with the subject’s visual perception.

\subsection{Dual-Constrained Diffusion Generation}

Building upon the semantic alignment and structural prediction modules, 
the final stage of SGDM integrates both priors to synthesize 
cognitively consistent visual representations. 
The goal is to generate images that faithfully reflect 
the subject’s perceived semantics and spatial layout 
while maintaining controllable and coherent reconstruction.

\subsubsection{Dual Conditioning Mechanism}

As illustrated in Fig.~\ref{fig:sgdm_pipeline}, 
the semantic embedding $z_I$ and predicted structural map $\mathbf{S}$ 
serve as dual conditioning inputs to the diffusion process. 
The semantic feature $z_I$ is injected into the SDXL-turbo U-Net 
via the \textit{IP-Adapter}~\cite{ye2023ipadapter} to ensure high-level semantic consistency, 
while the structural map $\mathbf{S}$ is integrated through \textit{ControlNet}~\cite{zhang2023controlnet} 
to constrain spatial geometry and preserve object layout. 
This dual-conditioning design effectively balances semantic alignment 
and structural fidelity during image synthesis.

\subsubsection{Generation Process}

Starting from Gaussian noise, 
the conditioned U-Net iteratively denoises through 2–4 steps to reconstruct the final image:
\[
\hat{x}_0 = \mathrm{U\text{-}Net}(z_I, \mathbf{S}, \epsilon_t)
\]
where $\epsilon_t$ denotes the predicted noise at time step $t$. 
By combining EEG-derived semantic and structural guidance, 
SGDM enables robust, semantically grounded, and spatially coherent 
EEG-to-vision reconstruction without requiring additional fine-tuning.

\section{Results}
\subsection{EEG Semantic Alignment: Integrating Multilevel Representations}
Achieving effective alignment between EEG signals and the semantic space of cognitive imagery is a critical prerequisite for ensuring accurate image generation. We first visualized the EEG embeddings before and after alignment across multiple representational levels to evaluate the effectiveness of the alignment process in integrating hierarchical representations and explaining individual variability.

\subsubsection{Group-Level EEG Alignment}
\begin{figure}[t]
    \centering
    \includegraphics[width=0.80\linewidth]{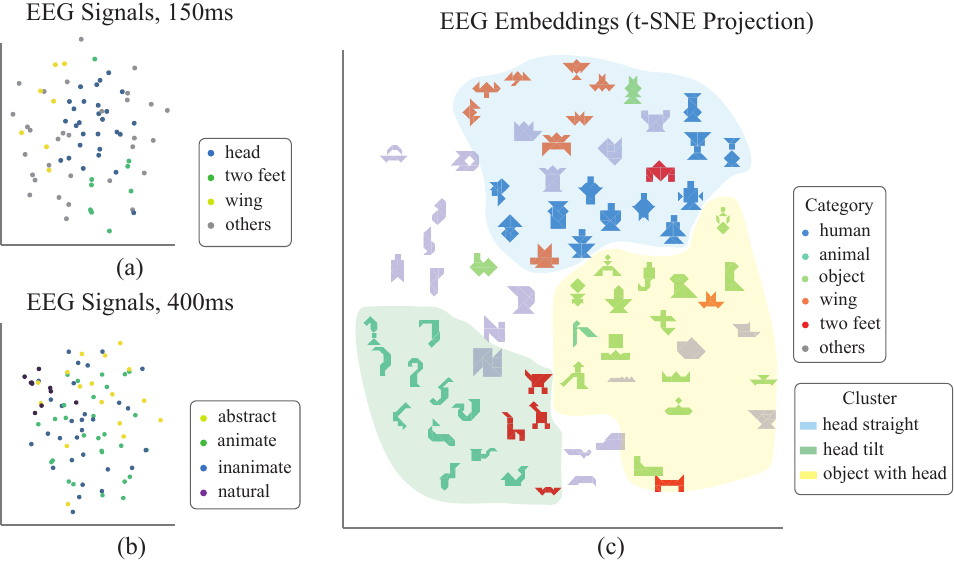}
    \caption{
    Comparison of EEG distributions before and after alignment.
    (a) Dimensionality-reduced visualization of pre-alignment EEG signals around 150 ms, highlighting local feature clustering (e.g., head, feet).
    (b) Dimensionality-reduced visualization of pre-alignment EEG signals around 400 ms, showing semantic clustering (e.g., animal vs. non-animal).
    (c) Dimensionality-reduced visualization of post-alignment EEG embeddings. Colors of tangrams indicate categories; broad color clusters reflect shared cross-category features.
    }
    \label{fig:r1}
\end{figure}

To preserve local neighborhood structures and reveal potential clustering patterns, we employed the t-distributed Stochastic Neighbor Embedding (t-SNE) algorithm~\cite{maaten2008visualizing} to project the EEG features before (Fig.~\ref{fig:r1}(a)(b)) and after (Fig.~\ref{fig:r1}(c)) alignment into a two-dimensional space. The results show that, before alignment, EEG signals exhibit distinct representational patterns in early (approximately 150 ms) and late (approximately 400 ms) temporal windows: early responses mainly capture local shape-related features (e.g., head, limbs, wings), while late responses reflect higher-level categorical distinctions (e.g., animal vs. non-animal). This “local-to-category” temporal progression is consistent with our previous representational analyses on the same dataset~\cite{lian2025multidimensional} and aligns with established findings from prior visual decoding studies~\cite{grootswagers2019representational,foster2017alpha}.

After alignment, the two-dimensional visualization of EEG embeddings revealed that the semantic groupings of the tangram images (e.g., human, animal, object, and structural groups characterized by wings or limbs) formed clearly separated clusters in the low-dimensional space. This result indicates that the EEG encoder successfully captured category-level semantic information while preserving local structural consistency. From the overall distribution (Fig.~\ref{fig:r1}(c)), three major clusters and their transitional regions can be distinguished, corresponding to cross-category shared local structural patterns (e.g., the upper-view head of biological figures, lateral head profiles, and salient parts of objects). Such cluster–category hierarchical correspondence demonstrates that the aligned EEG embeddings are capable of simultaneously integrating early local features and late semantic representations. The extraction and preservation of these multi-level representations lay the foundation for subsequent cognitive image reconstruction.

\subsubsection{Within-Subject EEG Alignment and Diversity}
\begin{figure}[t]
    \centering
    \includegraphics[width=0.8\linewidth]{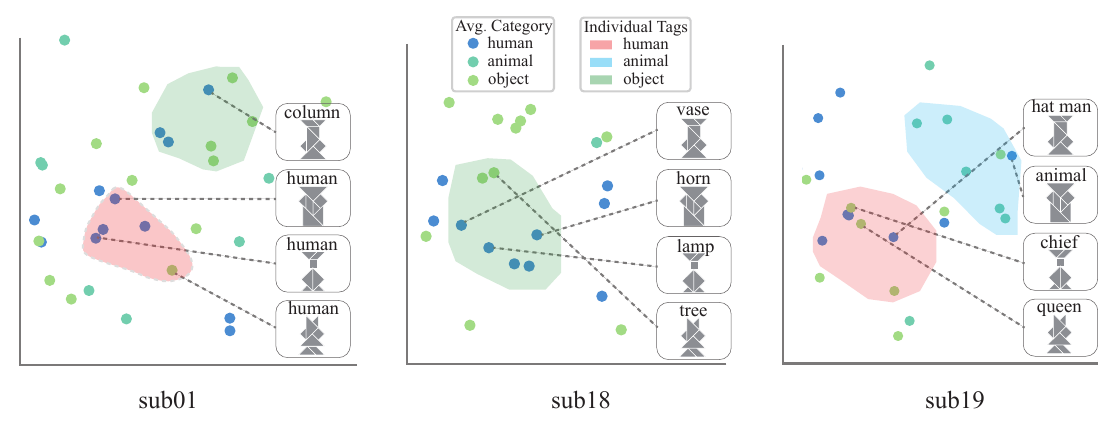}
    \caption{
        Subject-specific EEG embeddings after alignment and inter-subject semantic differences.
        From left to right are the aligned EEG embedding distributions of Sub01, Sub18, and Sub19. The panels on the right show semantic annotation differences for the same set of tangram figures across subjects. Scatter colors denote the cross-subject average semantic category of each EEG sample, while clustered color regions indicate subject-specific semantic patterns.
    }
    \label{fig:r2}
\end{figure}

To further verify whether the aligned EEG representations could retain subject-specific cognitive characteristics, we analyzed the embedding distributions across different participants using the Kilogram abstract visual object dataset. Fig.~\ref{fig:r2} illustrates the visualized EEG embeddings of three subjects (Sub01, Sub18, and Sub19). For each subject, four similar tangram images were selected, and their corresponding EEG embeddings and semantic annotations were labeled.

The results show notable inter-subject variability in semantic labeling. Sub01 labeled one sample as “column,” while the remaining three were annotated as “human.” Sub18 tended to interpret the same images as various objects, whereas Sub19 regarded three images as differently dressed humans and one as an animal. Despite these subjective semantic differences, the aligned EEG embeddings within each individual formed compact intra-class clusters and clearly separated inter-class distributions. This demonstrates that our cross-modal alignment method effectively integrates individual semantic representations while maintaining the structural consistency of the cognitive space.

These findings further validate the effectiveness of our alignment strategy from the perspective of individual specificity. The EEG encoder not only captures hierarchical local and categorical information but also preserves semantic differences across subjects, thereby enabling accurate and personalized cognitive image generation.
\subsection{Stage I: Structure Reconstruction from EEG}

Introducing structural priors into multilevel visual representations constitutes the core concept of our cognitive image reconstruction framework. The first question to address, therefore, is whether EEG signals contain sufficient structural information that can be explicitly predicted and extracted. In our previous Kilogram EEG experiment~\cite{lian2025multidimensional}, distinct structural types exhibited clear clustering patterns in representational similarity analysis (RSA) embeddings, while temporal generalization analysis further revealed differentiated activation and suppression of local features across participants. These results indicate that EEG signals indeed possess the potential to encode structural characteristics of visual objects.

\begin{figure}[t]
    \centering
    \includegraphics[width=0.80\linewidth]{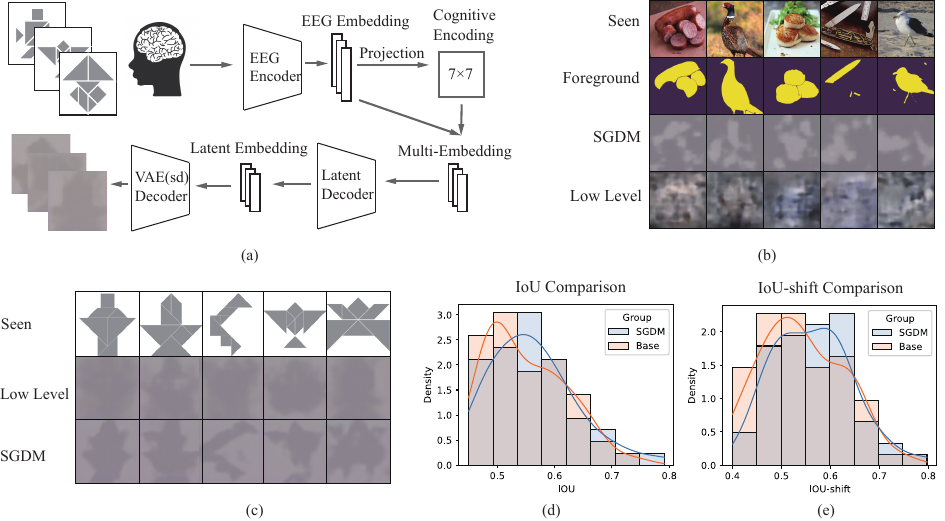}
    \caption{
        Pipeline, reconstruction results, and evaluation of structural information.
        (a) The generative model from EEG to structural information. The 7×7 cognitive code~\cite{lian2025multidimensional} serves as a key structural prior for supervision.
        (b) Comparison of structural reconstruction results between low-level feature extraction~\cite{li2024visual} and SGDM on the THINGS dataset.
        (c) Comparison of structural reconstruction results between low-level feature extraction and SGDM on the Kilogram dataset.
        (d) IoU evaluation of the reconstructed structures by the two methods on the Kilogram dataset.
        (e) Shift-IoU evaluation of the reconstructed structures by the two methods on the Kilogram dataset.
    }
    \label{fig:r3}
\end{figure}

Building upon these findings, we designed a structural information extraction and generation pipeline (Fig.~\ref{fig:r3}(a)). In this framework, the model first generates a structured image of the target foreground via a variational autoencoder (VAE), guided jointly by EEG embeddings and cognitive encodings~\cite{lian2025multidimensional}. The extracted structural information is then utilized to guide the subsequent diffusion-based cognitive image reconstruction. We conducted experiments and evaluations on both the natural image dataset THINGS and the abstract Kilogram dataset, comparing the proposed SGDM structural extraction method with existing low-level feature extraction approaches\cite{li2024visual}.

Fig.~\ref{fig:r3}(b) illustrates the structural generation results on the THINGS dataset. Here, the Base method for comparison is set as the previously proposed low-level feature extraction approach~\cite{li2024visual} that integrates hierarchical representations. The results show that the structural extraction stage reliably captures the global shape of target objects, providing a solid basis for subsequent structure-supervised learning. Compared with Base methods, SGDM produces structures that approximate the spatial layout of the foreground through compositional shape modeling, exhibiting higher interpretability and structural consistency.

Fig.~\ref{fig:r3}(c) presents the structural generation results on the Kilogram dataset. Given that Tangram images possess explicit geometric compositions and hierarchical relations, the model was trained under direct supervision of structural signals (cognitive encoding\cite{lian2025multidimensional}). The results demonstrate that existing low-level feature extraction methods generate blurred or distorted boundaries, deviating substantially from the ground truth. In contrast, SGDM more accurately reconstructs the block composition and spatial relationships of the stimuli. Quantitative results (Fig.~\ref{fig:r3}(d)(e)) show that SGDM achieves a higher mean Intersection-over-Union (IoU) between generated and target structures($p=0.013$), and significantly increases the proportion of high-quality structural reconstructions($p=0.008$).

Beyond generation quality, SGDM also exhibits superior training stability. Owing to the cognitively guided encoding supervision and the use of a deconvolutional architecture (see Supplementary data), SGDM effectively captures the key geometric features of Tangram objects and supports controllable structural editing, while mitigating the overfitting tendencies observed in low-level feature extraction Base model.

\subsection{Stage II: Cognitive Image Reconstruction with Structural Guidance}

In the image generation experiments, we further validated the capability of the SGDM model to perform cognitive image reconstruction under structural guidance. Fig.~\ref{fig:r4}(c) illustrates the overall pipeline, in which SGDM reconstructs realistic images conditioned on abstract Tangram stimuli, along with the corresponding evaluation pipeline. To handle the inherent uncertainty of Tangram annotations and to capture their abstract semantic characteristics, we proposed a similarity-based evaluation scheme for generated images. Specifically, we fine-tuned the CLIP model jointly on Tangram images and their semantic annotations, ensuring comparability between abstract stimuli and real visual images within a shared CLIP embedding space. This allowed us to compute the CLIP feature similarity between generated and target images, thereby quantifying the semantic consistency and cognitive plausibility of the generated outputs.

\subsubsection{Results and Evaluation of Cognitive Image Reconstruction}
\begin{figure}[t]
    \centering
    \includegraphics[width=0.85\linewidth]{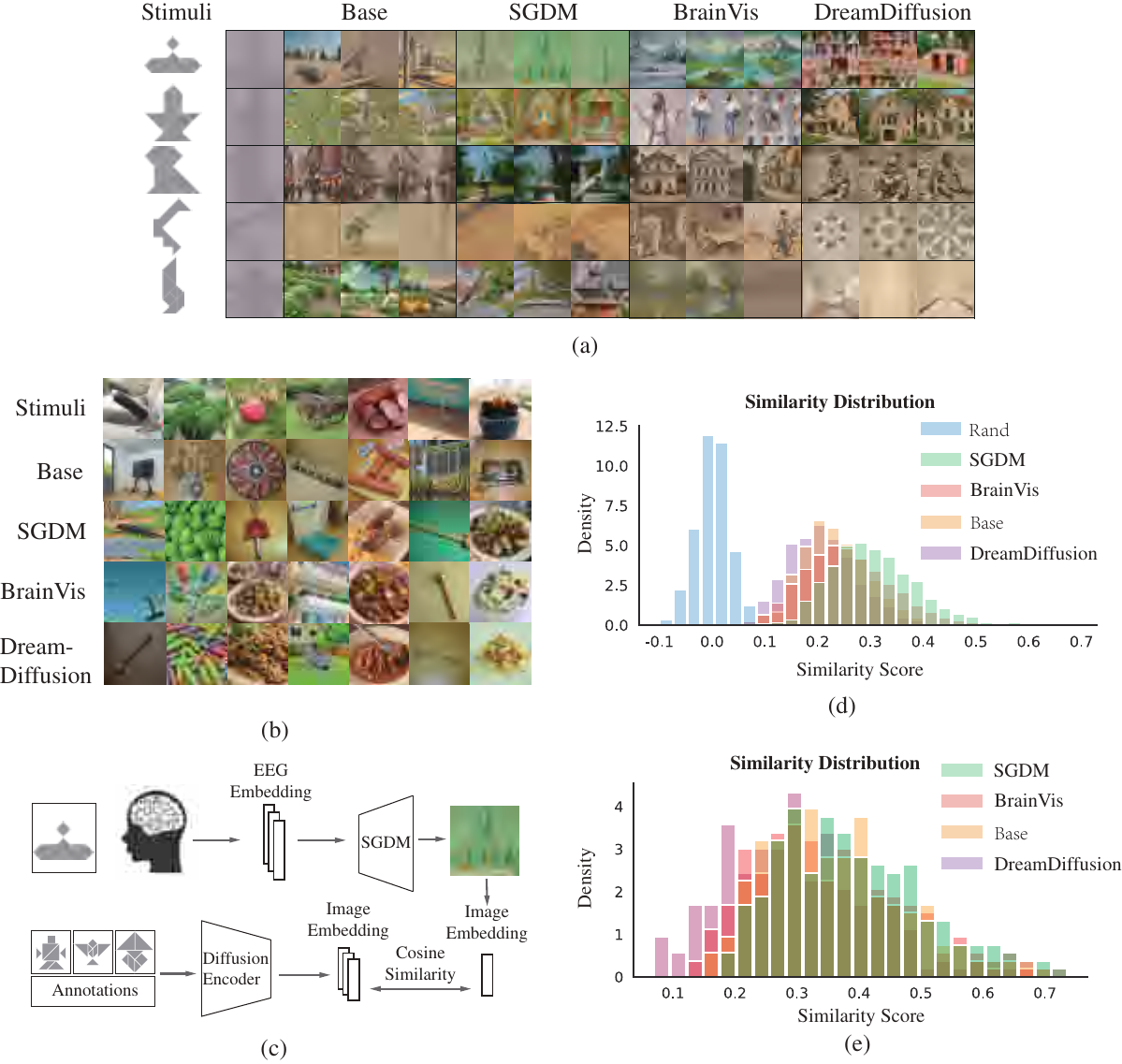}
    \caption{
        Performance and framework of cognitive image generation. 
        (a)Visual comparison of cognitive image generation results among SGDM and different reconstruction methods on the Kilogram dataset. 
        (b)Visual comparison of cognitive image generation results among SGDM and different reconstruction methods on the THINGS dataset. 
        (c)Pipeline of the SGDM framework for cognitive image generation, including the neural decoding stage and the evaluation procedure.
        (d)Similarity-based quantitative evaluation comparing SGDM with other representative methods on the Kilogram dataset. 
        (e) Similarity-based quantitative evaluation comparing SGDM with other representative methods on the THINGS dataset.
    }
    \label{fig:r4}
\end{figure}

Fig.~\ref{fig:r4}(a),(b) show the generation results on the THINGS and Kilogram (Tangram) datasets, respectively, in comparison with other representative generation model\cite{fu2025brainvis,Zhou_2024_CVPR}. For comparison, the Base method is still a model that leverages multi-level representations to guide the diffusion process ~\cite{li2024visual}. These examples were uniformly sampled from the results based on semantic categories, without any additional selection criteria.

On the THINGS dataset, it can be observed that both SGDM and the baseline methods are capable of generating images that exhibit a relatively high degree of semantic similarity to the viewed stimuli (e.g., food items, containers, elongated objects). However, SGDM demonstrates a clear advantage in capturing structural information of objects. This is particularly evident for visually complex structures—such as multiple stacked sausages, horizontally elongated rods, and densely arranged items—where SGDM more accurately reconstructs spatial layouts and geometric relationships.

In contrast, the reconstructions generated by BrainVis and DreamDiffusion exhibit less stable semantic representations. The core strength of the proposed SGDM framework lies in its brain-inspired dual constraints on structural geometry and semantic information. These results suggest that, while maintaining relatively accurate semantic predictions, the structural information pathway in SGDM effectively regularizes and guides the image reconstruction process.

On the Kilogram dataset, we evaluated both structural and semantic reconstruction performance. SGDM showed a striking advantage in structural reconstruction, producing segmented and spatially coherent object arrangements that closely matched the stimuli. In contrast, the Base model yielded blurred and structurally inconsistent results. At the semantic level, due to the inherently ambiguous categorical nature of Tangram stimuli, the Base model’s performance declined substantially. SGDM, however, generated semantically plausible images that corresponded meaningfully with participants’ annotations (e.g., “mountain,” “seated person,” “glass trophy”), demonstrating the model’s ability to generalize abstract cognitive semantics from EEG signals.

Fig.~\ref{fig:r4}(d),(e) present quantitative evaluations of cognitive image reconstruction. We computed the CLIP feature similarity between generated and viewed images for random generation, SGDM and other representative methods across both THINGS and Kilogram. Results show that all methods significantly outperform random generation, while SGDM consistently achieves higher mean similarity scores on Kilogram ($p=0.021$) and THINGS ($p=0.037$), confirming that incorporating structural priors leads to more accurate and cognitively consistent reconstructions.

\begin{table}[htbp]
    \centering
    \caption{Comparison of generation metrics on the Kilogram and THINGS}
    \label{tab:gen_metrics}
    \begin{tabular}{clcccc}
    \hline
    \textbf{Dataset} & \textbf{Model} & \textbf{SSIM \textuparrow} & \textbf{IS \textuparrow} & \textbf{SwAV FID \textdownarrow} & \textbf{CLIP Score \textuparrow} \\
    \hline
    \multirow{4}{*}{Tangram} 
      & Base & 0.198 & 7.070 & 49.979 & 0.222 \\
      & BrainVis & 0.232 & 7.497 & 47.631 & 0.331 \\
      & DreamDiffusion & 0.218 & 6.380 & 56.081 & 0.209 \\
      & SGDM & \textbf{0.322} & 6.208 & \textbf{38.999} & \textbf{0.396} \\
    \hline
    \multirow{4}{*}{THINGS} 
      & Base & 0.250 & 9.088 & 24.408 & 0.363 \\
      & BrainVis & 0.247 & 8.847 & 25.113 & 0.351 \\
      & DreamDiffusion & 0.245 & 8.594 & 27.637 & 0.307 \\
      & SGDM & \textbf{0.279} & 8.682 & \textbf{23.979} & \textbf{0.387} \\
    \hline
    \end{tabular}
\end{table}

When evaluating cognitive image generation, in addition to the CLIP similarity metric, we employed the low-level feature metric Structural Similarity Index Measure(SSIM)~\cite{wang2004image} to assess pixel-level consistency, the Inception Score(IS)~\cite{salimans2016improved} metric to evaluate image realism, and the Swapping Assignments between Views(SwAV) FID~\cite{caron2020unsupervised} metric to measure overall representation distance.

Building upon a systematic evaluation framework that spans from low-level visual features to high-level semantic consistency, Table~\ref{tab:gen_metrics} summarizes the image reconstruction performance of different algorithms across the two datasets. The four metrics, listed from left to right, respectively assess reconstruction quality in terms of low-level features, diversity, high-level semantics, and overall semantic alignment.

In comparison with other representative methods, the proposed SGDM achieves the best performance on SSIM, SwAV FID, and CLIP Score across both the Tangram and THINGS datasets. However, in terms of generative diversity, as measured by the Inception Score (IS), SGDM performs lower than most competing approaches. This behavior is consistent with the design principle of SGDM. By imposing dual constraints on the image generation process, SGDM reduces a certain degree of diversity while improving reconstruction fidelity.
Importantly, as illustrated by the qualitative results in Fig.~\ref{fig:r4}, this reduction in diversity does not significantly compromise the perceptual quality of the reconstructed images. 

Furthermore, among all evaluation metrics, SGDM exhibits the largest improvement in SSIM, indicating that the incorporation of structural priors via the structural representation pathway effectively enhances the reconstruction of low-level visual features.

\subsubsection{Cross-Subject Variability and Cognitive Diversity in Reconstruction}
\begin{figure}[t]
    \centering
    \includegraphics[width=0.80\linewidth]{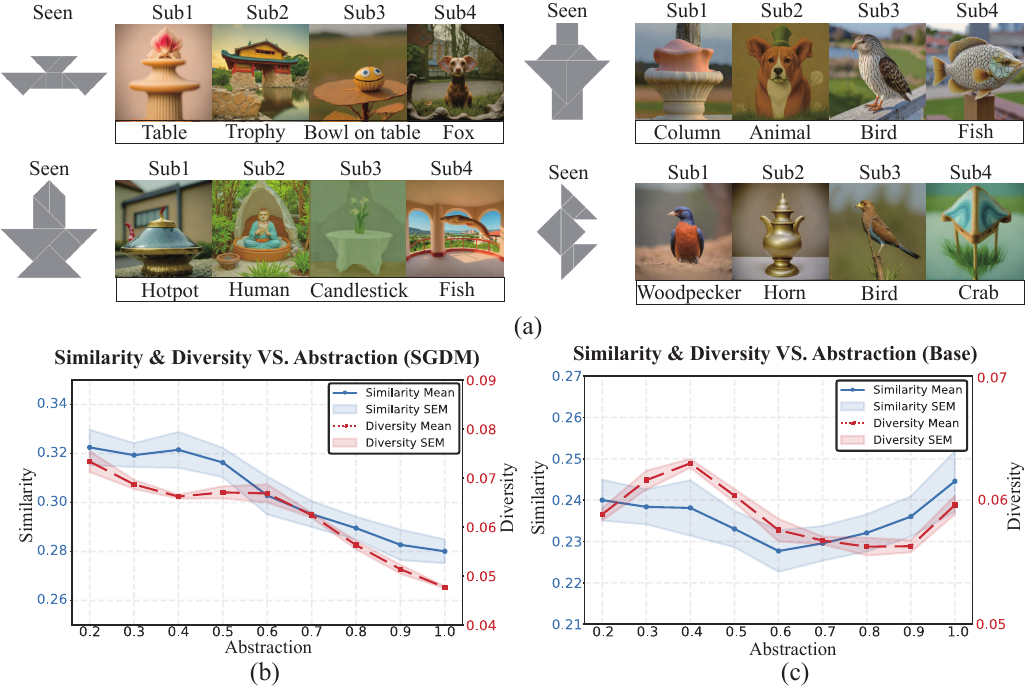}
    \caption{
        Subject-specific cognitive image generation and the effect of image abstraction parameters on reconstruction.
        (a) Cognitive reconstructions and corresponding annotations of the same tangram figure across different subjects.
        (b) Similarity and diversity trends of cognitive reconstructions at different abstraction levels using the SGDM method.
        (c) Similarity and diversity trends of cognitive reconstructions at different abstraction levels using the Base method.
    }
    \label{fig:r5}
\end{figure}

A particularly noteworthy finding lies in the cognitive diversity of Tangram stimuli. The same Tangram configuration can elicit distinct semantic interpretations across participants, posing a challenging test for EEG-based reconstruction models to extract stable yet individualized representations. Fig.~\ref{fig:r5}(a) illustrates SGDM’s reconstructions of the same Tangram stimulus across different participants, together with their corresponding semantic annotations. SGDM successfully generated semantically diverse real-world images—such as “hotpot,” “seated person,” “table,” and “candlestick”—while maintaining reasonable visual correspondence to the original Tangram layout. This demonstrates SGDM’s capacity to capture participant-specific semantic representations from EEG.

Furthermore, we quantitatively analyzed the relationship between image similarity and stimulus abstraction level (Detailed information is provided in Supplementary data). As shown in Fig.~\ref{fig:r5}(b), within the SGDM model, similarity between generated and stimulus images decreases as the abstraction level increases(Pearson’s $r=-0.97$,$p<0.001$), consistent with the intuitive expectation that higher abstraction imposes greater reconstruction difficulty. In contrast, the Base model (Fig.~\ref{fig:r5}(c)) exhibited no significant trend(Pearson’s $r=-0.11$,$p=0.73$), consistent with its limited generative performance on the Tangram dataset. Collectively, these findings demonstrate that SGDM achieves, for the first time, the modeling and reconstruction of individual cognitive diversity under abstract visual conditions, showing strong adaptability across subjects and abstract semantic contexts in EEG-based cognitive image generation.

\begin{figure}[htpb]
    \centering
    \includegraphics[width=0.50\linewidth]{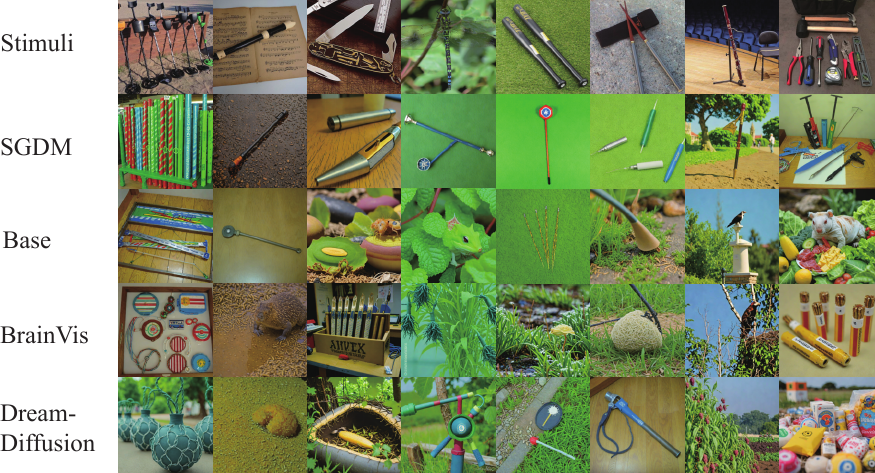}
    \caption{
        Comparative reconstruction performance on natural images with cognitive ambiguity.
    }
    \label{fig:rx}
\end{figure}

The proposed SGDM framework demonstrates cognitively consistent image reconstruction not only on Tangram stimuli but also on natural images. To better evaluate subjective cognitive alignment, we focus on natural images with cognitive ambiguity, where semantic interpretations are inherently uncertain.

As shown in Fig.~\ref{fig:rx}, baseline models, BrainVis, and DreamDiffusion exhibit unstable and often incoherent reconstructions under such conditions, frequently failing to align with the actual stimuli (e.g., misinterpreting a minesweeper device as “landmines” or producing disorganized outputs for cluttered tools).

In contrast, SGDM captures salient cognitive structures despite semantic ambiguity. For example, even when participants cannot recognize a minesweeper device, they reliably perceive “multiple vertically arranged components,” which is reflected in SGDM’s reconstruction. Similarly, for unfamiliar objects such as orchestral instruments, SGDM preserves the perception of “elongated vertical structure.” These results indicate that SGDM effectively leverages structural cues to achieve reconstruction aligned with subjective cognition, even when explicit semantics are unclear.

\subsection{Spatio-Temporal Analysis of EEG-Based Cognitive Image Reconstruction}

To further explore the temporal and spatial characteristics of the SGDM reconstruction framework, we conducted a series of spatiotemporal experiments to evaluate how different time windows and cortical regions contribute to reconstruction performance.

\subsubsection{Temporal Analysis}
\begin{figure}[t]
    \centering
    \includegraphics[width=0.80\linewidth]{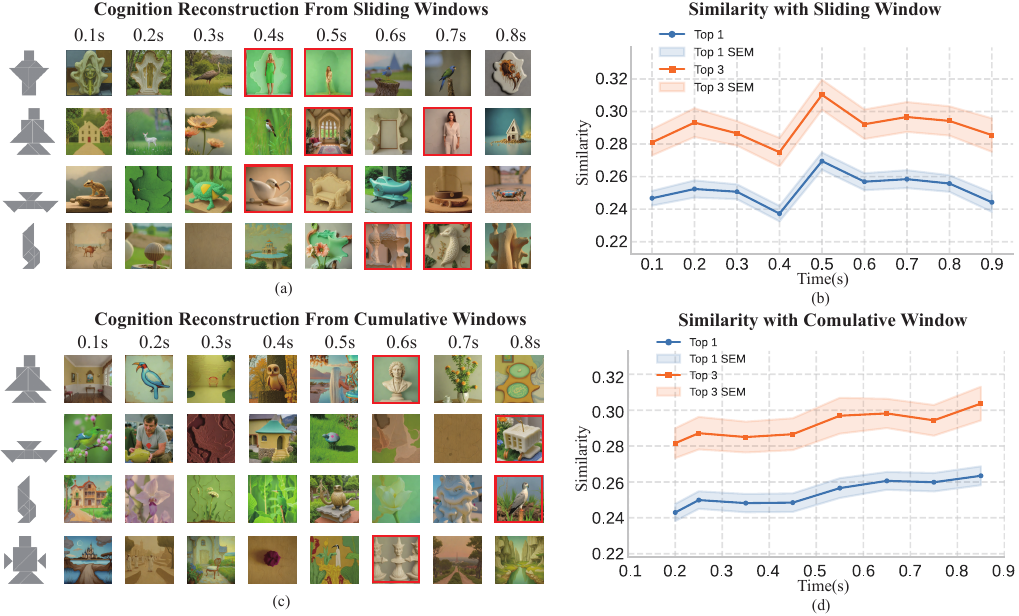}
    \caption{
        SGDM-based cognitive reconstruction and evaluation across different EEG time windows.
        (a) Cognitive image reconstruction results under sliding time windows. Red boxes highlight reconstructed samples that accurately match the original stimuli.
        (b) Similarity curves of cognitive reconstructions with varying sliding time windows.
        (c) Cognitive image reconstruction results under cumulative time windows.
        (d) Similarity curves of cognitive reconstructions with varying cumulative time windows.
    }
    \label{fig:r6}
\end{figure}

For temporal analysis, we employed two time-windowing strategies: cumulative windows and sliding windows. In the cumulative window experiment, EEG segments were selected from the stimulus onset (0 ms) up to varying endpoints, progressively extending the temporal range. In contrast, the sliding window experiment used a fixed window width of 200 ms, shifting this window across the entire post-stimulus period. Reconstruction quality was assessed by computing the CLIP feature similarity between generated images and their corresponding target stimuli.

As shown in Fig.~\ref{fig:r6}(c)(d), in the cumulative window condition, the highest reconstruction similarity occurred within the 0–500ms time range; both shorter and longer windows led to performance degradation. In the sliding window condition (Fig.~\ref{fig:r6}(a)(b)), reconstruction similarity increased steadily before peaking around 500 ms, followed by minor fluctuations. Collectively, these results suggest that the 400–600 ms interval contributes most substantially to the reconstruction of abstract Tangram stimuli. This temporal range is slightly later than the typical peak window observed for natural image decoding (200–300 ms)~\cite{grootswagers2024mapping}, but aligns well with prior findings on abstract visual processing~\cite{lian2025multidimensional}, indicating the additional cognitive load required for semantic abstraction.

\subsubsection{Spatial Analysis}
\begin{figure}[t]
    \centering
    \includegraphics[width=0.65\linewidth]{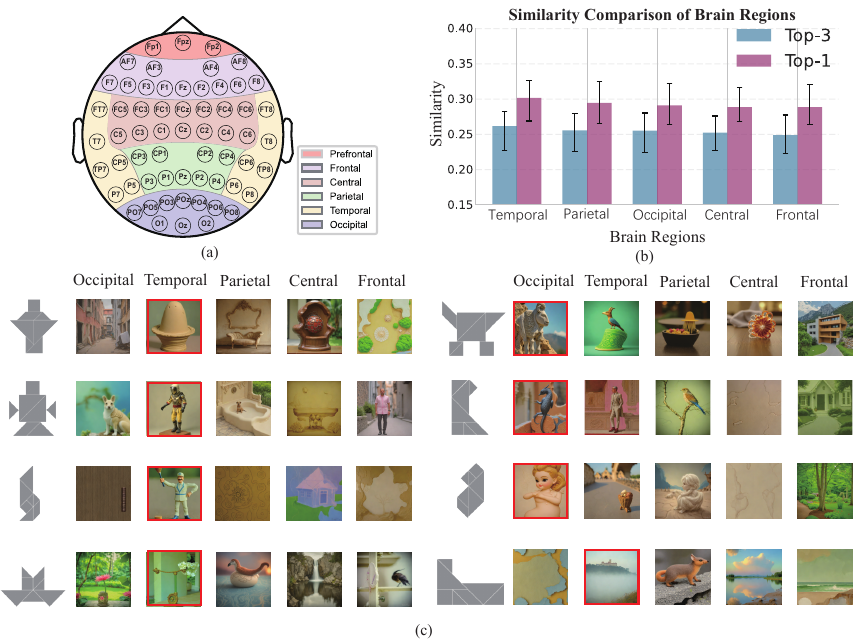}
    \caption{
        SGDM-based cognitive reconstruction and evaluation across different brain regions.
        (a) EEG electrode distributions corresponding to different brain regions.
        (b) Bar chart of cognitive reconstruction similarity across 5 brain regions. Error bars represent the standard error of the mean (SEM).
        (c) Cognitive image reconstruction results based on EEG signals from 5 brain regions under the same tangram stimulus condition. Red boxes highlight reconstructed samples that accurately match the original stimuli.
    }
    \label{fig:r7}
\end{figure}

To further assess the spatial contributions of different brain regions, we divided EEG channels into five major brain areas: frontal, parietal, central, temporal, and occipital (Fig.~\ref{fig:r7}(a)). Using the same CLIP-based similarity metric, we evaluated the reconstruction performance for each region separately. As shown in Fig.~\ref{fig:r7}(b)(c), the temporal region exhibited the highest reconstruction similarity ($p=0.043$), followed by the parietal and occipital regions.

This finding contrasts with previous results from natural image decoding studies, where the occipital cortex typically achieves the highest decoding accuracy~\cite{li2024visual}, consistent with its role as the core region for early visual processing. In the Tangram condition, however, low-level visual differences among stimuli are minimal, and the primary distinctions arise from higher-level semantic and cognitive processing. Consequently, occipital signals contribute less to reconstruction, while activity from the temporal and parietal cortices—which are associated with semantic integration, mental imagery, and cognitive control—plays a dominant role.

In summary, the temporal analysis reveals that abstract visual reconstruction relies heavily on later processing stages (400–600 ms), while the spatial analysis highlights the critical involvement of temporal and parietal regions during these stages. Together, these findings indicate that SGDM’s success in reconstructing abstract visual stimuli depends on the extraction of structured representations from late cognitive processing, rather than early perceptual signals alone.

\subsection{Ablation on Structural and Generative Components}
To analyze the contribution of SGDM’s core components and key parameters to cognitive image reconstruction performance, we conducted a series of ablation experiments. The study consisted of two parts: (1) evaluating the roles of the structure module and structural information during the structure prediction stage, and (2) analyzing the influence of the generation control coefficient during the conditional generation stage.

\subsubsection{Structure-Guided Module Ablation}

Structural guidance serves as the central mechanism of the SGDM framework, and validating the effectiveness of structural priors is essential. We employed Intersection over Union (IoU) and Shifted-IoU as quantitative metrics for evaluating structure generation, and compared three configurations: No Component — removing the structural processing module; Zero Information — retaining the module but setting the input structural information to zero; Structure Guided — the full SGDM model incorporating explicit structural priors.

\begin{table}[htbp]
    \centering
    \caption{Ablation study of reconstruction accuracy across different datasets}
    \label{tab1}
    \begin{tabular}{llcc}
    \hline
    \textbf{Dataset} & \textbf{Method} & \textbf{IoU} & \textbf{IoU shift} \\
    \hline
    \multirow{3}{*}{Kilogram} 
      & No component & 0.492 & 0.519 \\
      & Zero information & 0.494 & 0.520  \\
      & SGDM & \textbf{0.517} & \textbf{0.543} \\
    \hline
    \multirow{3}{*}{THINGS} 
      & No component & 0.432 & 0.472 \\
      & Zero information & 0.459 & 0.481  \\
      & SGDM & \textbf{0.541} & \textbf{0.576} \\
    \hline
    \end{tabular}
\end{table}

As shown in Tab.~\ref{tab1}, the inclusion of structural guidance significantly improved structural prediction accuracy on both the Kilogram (Tangram) and THINGS datasets. The performance gain was particularly prominent on the THINGS dataset (IoU on THINGS increased by 0.109, on Kilogram increased by 0.025), indicating that in complex natural scenes, structural priors play a crucial role in ensuring generation stability and spatial consistency.

In the second stage of SGDM, final cognitive image reconstruction is performed under the guidance of the predicted structural information. Therefore, we further evaluated the effect of the generation control coefficient, which determines the relative weighting between structural and semantic constraints. Five control values within the range 0-1 were tested, using three evaluation metrics: Structural Similarity Index (SSIM) for low-level feature alignment, Inception Score (IS) for image realism, and SwAV-FID for joint assessment of semantic and structural consistency.

\subsubsection{Generation Control Parameter Ablation}
\begin{table}[htbp]
    \centering
    \caption{Comparison of generation metrics under different control rates on the Kilogram}
    \label{tab2}
    \begin{tabular}{cccc}
    \hline
    \textbf{Control Rate} & \textbf{SSIM \textuparrow} & \textbf{IS \textuparrow} & \textbf{SwAV FID \textdownarrow} \\
    \hline
      0.00 & 0.253 & 6.864 & 39.913 \\
      0.25 & 0.278 & 6.829 & 38.891 \\
      0.50 & 0.308 & 6.298 & \textbf{38.424} \\
      0.75 & 0.345 & 5.778 & 40.162 \\
      1.00 & 0.373 & 4.980 & 43.162 \\
    \hline
    \end{tabular}
\end{table}

As summarized in Tab.~\ref{tab2}, increasing the control coefficient led to a monotonic rise in SSIM but a decline in IS, revealing a trade-off between fine-grained structural fidelity and perceptual realism. The comprehensive SwAV-FID metric first decreased and then increased, achieving its minimum around 0.5, suggesting an optimal balance point between low-level and high-level representations. Accordingly, we selected 0.5–0.6 as the optimal control range for subsequent experiments.

Overall, these results demonstrate that structural priors are indispensable for multi-level visual reconstruction, and that a moderate control strength effectively balances structural precision and semantic coherence. This confirms the architectural soundness and robustness of the SGDM framework in cognitive image generation.

\section{Discussion}
Compared with previous visual decoding approaches that primarily rely on natural images and categorical labels~\cite{takagi2023high,li2024visual,cheng2025fine}, the proposed Structure-Guided Diffusion Model (SGDM) explicitly models the hierarchical relationship between structural information and semantic representations through a two-stage design. By incorporating both natural and abstract image datasets, SGDM achieves structured and controllable generation that bridges the perceptual and cognitive levels.
In the first-stage structure generation, SGDM effectively integrates the spatiotemporal features contained in EEG signals and captures both the global configuration and local composition of visual objects through the structure prediction module. In the second-stage conditional generation, the experimental results demonstrate that SGDM significantly improves structural consistency and cross-subject generalization. Particularly in the abstract visual task using Tangram stimuli, the model not only reconstructs the geometric layout of objects with high fidelity but also reflects individual differences in subjective semantic interpretation, showing sensitivity to personalized cognitive representations. These findings collectively verify the critical role of structural information in cognitive-level image generation and demonstrate the feasibility of controllable visual cognition modeling at the EEG level.

From the perspective of neural mechanisms, the spatiotemporal analyses reveal that abstract visual reconstruction mainly depends on EEG activity within the 400–600 ms post-stimulus interval, with substantial contributions from the temporal and parietal regions. This time window occurs later than the early visual responses typically observed for natural image recognition (200–300 ms)~\cite{grootswagers2019representational}, suggesting that abstract visual processing involves deeper semantic integration and higher-order cognitive engagement. Furthermore, the structure-guided constraint introduced in SGDM aligns closely with the theory of distributed hierarchical processing in the brain~\cite{felleman1991distributed}. By jointly incorporating structural and semantic constraints during the diffusion generation stage, SGDM partially emulates the human brain’s integration process from local shape encoding to global semantic composition. This structure-guided generation strategy endows the model with a degree of neural interpretability and provides a novel computational pathway for understanding the representational mechanisms underlying visual cognition.

Despite these promising results, several limitations remain. First, the current experiments are based on a relatively small dataset of abstract graphical stimuli, and the limited sample size and semantic diversity may constrain the model’s generalization ability. Second, the low spatial resolution of EEG poses challenges for precisely capturing fine-grained structural features. Future studies may address these issues by (1) validating the cross-modal consistency of SGDM on larger and multimodal datasets such as EEG–fMRI or EEG–fNIRS; (2) exploring online learning and few-shot individual adaptation strategies to enable real-time cognitive image generation; and (3) investigating potential applications of SGDM in clinical BCI, consciousness assessment, and cognitive rehabilitation.

In summary, the proposed SGDM framework demonstrates the feasibility of extracting and utilizing structural information from EEG signals for cognitive image reconstruction. By integrating structure guidance, semantic alignment, and conditional diffusion generation, SGDM achieves a unified balance between high-level semantic coherence and low-level spatial controllability. This work not only advances the frontier of EEG-based perceptual and cognitive image reconstruction but also provides a novel modeling and analytical framework for understanding the hierarchical processing mechanisms of human visual cognition.

\section{Conclusion}
This study proposes a novel EEG-based visual cognition reconstruction framework, termed the Structure-Guided Diffusion Model (SGDM). The model introduces structural guidance and dual semantic constraints into the generative mechanism, achieving highly consistent reconstruction of cognitive images from EEG signals through explicit structure prediction, cross-modal semantic alignment, and controllable diffusion generation. Experimental results demonstrate that SGDM significantly enhances structural fidelity, semantic coherence, and cross-subject generalization under abstract visual stimulation, while revealing the critical role of late-stage EEG activity and the temporal–parietal regions in the formation of structured cognitive representations.

The development of SGDM not only verifies the existence of hierarchically structured information within EEG signals that can be explicitly modeled but also provides new insights into the neural encoding mechanisms of visual cognition. By integrating the hierarchical processing theory of neuroscience with generative diffusion modeling, this study highlights the great potential of cross-modal information fusion in advancing cognitive image reconstruction.

Looking forward, the SGDM framework can be extended to multimodal neural signals (e.g., fMRI and MEG) and cross-task decoding scenarios to enable higher-resolution and semantically richer reconstructions. The approach also shows potential for applications in BCIs, psychological assessment, and neurorehabilitation. Overall, this work provides a framework for decoding both externally driven perception and internally generated subjective cognitive representations from neural signals, supporting BCIs with increased degrees of freedom for intention decoding and more flexible brain-to-machine communication.

\section*{About Preprint}
This is the version of the article before peer review or editing, as submitted by an author to Journal of Neural Engineering.

\bibliographystyle{unsrtnat}
\bibliography{references}  






\end{document}